\documentclass[conference]{IEEEtran}
\IEEEoverridecommandlockouts
\usepackage{amsmath,amssymb,amsfonts}
\usepackage{algorithmic}
\usepackage{graphicx}
\usepackage{textcomp}
\usepackage[linesnumbered,ruled,vlined]{algorithm2e}

\usepackage[table]{xcolor}
\usepackage{booktabs}
\usepackage{xcolor}
\usepackage{orcidlink}
\usepackage[export]{adjustbox}
\usepackage[backend=biber]{biblatex}
\usepackage{multirow}

\def\BibTeX{{\rm B\kern-.05em{\sc i\kern-.025em b}\kern-.08em
    T\kern-.1667em\lower.7ex\hbox{E}\kern-.125emX}}
\begin{document}

\title{TF-SASM: Training-free Spatial-aware Sparse Memory for Multi-object Tracking}

\author{\IEEEauthorblockN{Thuc Nguyen-Quang \orcidlink{0000-0003-2523-8851}}
\IEEEauthorblockA{
\textit{Software Engineering Lab \&} \\
\textit{John von Neumann Institute}\\
\textit{University of Science, VNU-HCM} \\
\textit{Viet Nam National University, Ho Chi Minh City} \\}
\and
\IEEEauthorblockN{Minh-Triet Tran \orcidlink{0000-0003-3046-3041}}
\IEEEauthorblockA{\textit{Faculty of Information Technology} \\
\textit{Software Engineering Lab \& John von Neumann Institute}\\
\textit{University of Science, VNU-HCM} \\
\textit{Viet Nam National University, Ho Chi Minh City}}}

\maketitle

\begin{abstract}
    Multi-object tracking (MOT) in computer vision remains a significant challenge, requiring precise localization and continuous tracking of multiple objects in video sequences. The emergence of data sets that emphasize robust reidentification, such as DanceTrack, has highlighted the need for effective solutions. While memory-based approaches have shown promise, they often suffer from high computational complexity and memory usage due to storing feature at every single frame. In this paper, we propose a novel memory-based approach that selectively stores critical features based on object motion and overlapping awareness, aiming to enhance efficiency while minimizing redundancy. As a result, our method not only store longer temporal information with limited number of stored features in the memory, but also diversify states of a particular object to enhance the association performance. Our approach significantly improves over MOTRv2 in the DanceTrack test set, demonstrating a gain of 2.0\% AssA score and 2.1\% in IDF1 score.
\end{abstract}


\section{Introduction}
\label{sec:intro}
In computer vision, tackling multi-object tracking (MOT) poses a significant challenge. This task involves precisely locating and tracking the temporal continuity of multiple objects within a video sequence \cite{milan2016mot16, dendorfer2020motchallenge}. The resultant trajectories of these objects are invaluable for various applications, spanning from action recognition to behavior analysis. The critical challenges of the MOT problem include occlusion, target rotation, target deformation, blurring, fast-motion objects, and the complexities introduced by camera motion \cite{deepsurvey}.

Historically, research efforts in Multiple Object Tracking (MOT) have primarily pursued two distinct directions: tracking-by-detection \cite{sort, deepsort} methodologies and end-to-end models \cite{centertrack, tracktor, memot, memotr}. Tracking-by-detection methods typically rely on object detectors to identify individual objects within each frame, followed by the application of filtering techniques like the Kalman filter to establish correspondence between bounding boxes across the video sequence \cite{sort, ocsort, bytetrack}. Some innovations in this realm have focused on enhancing reidentification by leveraging deep features \cite{deepsort, deepocsort, qdtrack}. In contrast, end-to-end models represent a more recent paradigm shift, consolidating both detection and association tasks within a unified deep-learning framework. Initially, earlier work often used CNN-based architecture, followed by the propagation of detected objects into subsequent frames \cite{tracktor, centertrack}. However, with the introduction of the DETR (DEtection TRansformer) model \cite{detr}, which leverages the Transformer architecture \cite{transformer} and stands as a state-of-the-art detector at that time, attention-based tracking approaches \cite{trackformer, transtrack, motr}, a subtype of end-to-end models, have emerged, demonstrating promising results in various tracking scenarios.

In recent years, the emergence of datasets like SportsMOT \cite{sportsmot} and DanceTrack \cite{dancetrack} has highlighted the critical need for robust reidentification methods, especially in scenarios featuring visually similar objects. While memory-based approaches \cite{memot, memotr} have shown promise in addressing these challenges, they often suffer from high computational complexity and memory usage, primarily due to storing all features for every frame. This redundancy can be further exacerbated by minor object movements between frames in high-fps video.

Our proposed method offers a streamlined solution: a sparse memory that selectively stores critical features based on object motion and overlapping awareness. This approach leverages the intuition that an object's degree of deformation correlates directly with its movement, thereby minimizing redundancy and enhancing long-temporal information. To validate our hypothesis, we incorporate our memory mechanism into the MOTRv2 \cite{motrv2} model due to its inherent flexibility. MOTRv2 represents a novel hybrid approach, combining tracking-by-attention and tracking-by-detection methodologies, thus harnessing the strengths of robust detectors and a transformer-like architecture for tracking tasks. This architecture is adaptable across diverse scenarios due to its separation of detector and tracker components. By extending this flexibility, we introduce a novel training-free memory mechanism for object tracking, thereby enhancing the association performance of the model. 

In summary, our contributions are as follows:
\begin{itemize}
    \item Introduce a Spatial-aware Sparse Memory, which selectively captures essential features based on object motion.
    \item Propose an Overlapping-aware Feature Selector, which effectively reduces noise from overlapping.
    \item Demonstrate the effectiveness of our proposed memory through improvements in key association metrics (2.0\% AssA score and 2.1\% IDF1 score) comparing with MOTRv2 on the DanceTrack test set, validating its impact on the model's ability to re-identify objects.
\end{itemize}

\section{Related work}
\label{sec:rel_work}

\subsection{Tracking-by-detection}
\label{sec:rw_tbd}
Previous methods in multi-object tracking (MOT) typically involved training a detector to locate objects in each frame individually, followed by linking these detections across the video sequence. One effective tracking method is shown in the SORT \cite{sort} algorithm. SORT predicts where an object will go next using a Kalman filter and then uses the Hungarian algorithm to link these predictions with new detections. DeepSORT \cite{deepsort} builds on SORT by adding more detailed object feature matching, which helps in identifying objects again after they disappear. Another notable approach, BYTETrack \cite{bytetrack}, handles situations where detections might have low confidence due to obstacles or blurriness. BYTETrack deals with these uncertain detections by associating them in a second pass, which reduces cases of wrongly identifying objects as absent. OC-SORT \cite{ocsort} introduces an innovative update mechanism for the Kalman filter: when an object reappears after being lost, the filter's parameters are adjusted based on the new observation. Deep OC-SORT \cite{deepocsort} extends this idea by also updating deep features adaptively for object reidentification, guided by the confidence scores of detections.

\subsection{End-to-end Approaches}
\label{sec:rw_e2e}

End-to-end object tracking models utilize two main approaches: tracking-by-regression and tracking-by-attention. Tracking-by-regression propagates object detections from one frame to the next, bypassing traditional matching methods. For example, Tracktor \cite{tracktor} uses a regression head to determine object existence in subsequent frames, while CenterTrack \cite{centertrack} predicts detections based on object centers from previous frames. With the introduction of DETR \cite{detr}, a new approach emerged, integrating detection and tracking using Transformer architecture. TrackFormer \cite{trackformer} builds on this, using an attention mechanism in the Transformer to treat objects from previous frames as queries for subsequent frames, enhancing tracking effectiveness. Similarly, MOTR \cite{motr} extends this concept by aggregating track queries across multiple frames. TransTrack \cite{transtrack} employs an IoU matching mechanism instead of a regression head for determining new object appearances or tracking previous ones.

\subsection{Hybrid Approaches and Memory-based Approaches}

Although tracking-by-attention methods are effective, they struggle with adapting to diverse object classes, requiring simultaneous training for detection and tracking. MOTRv2 \cite{motrv2} addresses this by combining tracking-by-detection and tracking-by-attention, using a pretrained detector and an attention mechanism to propagate objects across frames. Studies have shown that memory mechanisms enhance reidentification \cite{memotr, memot} in MOT. Unlike SOT, which deals with fewer objects, MOT involves more objects, making memory management challenging. Therefore, instead of storing object features at each frame, we propose a novel approach for memory, which stores object sparsely.

\section{Method}
\begin{figure}[t]
    \centering
    \includegraphics[width=\linewidth]{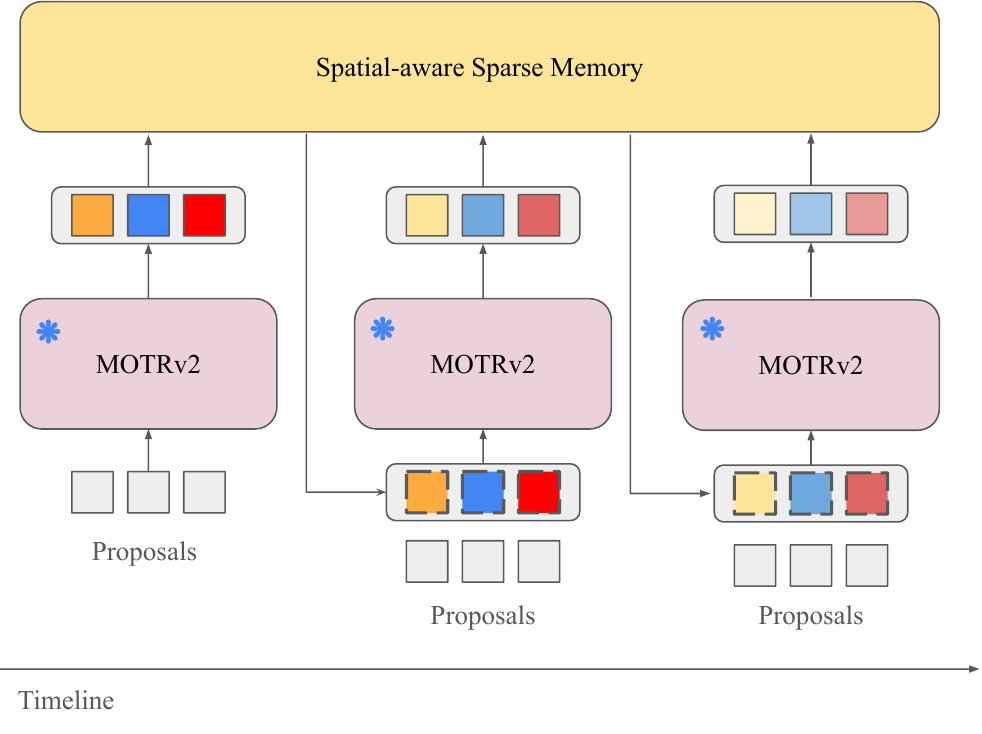}
    \caption{Our proposed method seamlessly integrates into the MOTRv2 \cite{motrv2} model as a training-free module, preserving MOTRv2's flexibility for various applications. As depicted in the figure, the pipeline takes proposals generated by YOLOX \cite{yolox2021} and previously tracked objects (detailed in Section \ref{sec:method_revisit}) as input for MOTRv2. The Spatial-aware Sparse Memory (SASM), further explained in Figure \ref{fig:SASMmot}, then processes MOTRv2's outputs.}
    \label{fig:overall}
\end{figure}
In this section, we present our method, which builds on the MOTRv2 \cite{motrv2} model known for its flexibility and superior tracking capabilities. We enhance MOTRv2 by introducing a memory module to improve reidentification performance. As shown in Figure \ref{fig:overall}, our memory module filters the object embeddings produced by MOTRv2 in each frame to retain the most important features. These selected features are then propagated to the next frame to aid in object localization. The filtering process includes two modules: Spatial-Aware Sparse Memory (SASM) and Overlapping-Aware Feature Selector (OFS), detailed in Sections \ref{sec:method_SASM} and \ref{sec:method_ofs} respectively.

\subsection{Revisiting MOTRv2}
\label{sec:method_revisit}
Y Zhang et. al. propose integrating the YOLOX \cite{yolox2021} object detector to provide detection priors for MOTR \cite{motr}. By providing detection priors, YOLOX reduces the burden on the tracker, which would otherwise be responsible for handling both detection and tracking tasks.

\subsubsection{Proposal query generation}
In this approach, for each frame $t$, YOLOX generates $N_t$ proposals $P$, each characterized by center coordinates $(x_t, y_t)$, height $h_t$, width $w_t$, and confidence score $s_t$. Initially, a query mechanism similar to the DAB-DETR \cite{liu2022dabdetr} architecture is employed for the first frame The confidence scores are also integrated into the proposal queries $\mathbf{q}_t$ using sine-cosine positional encoding.

\subsubsection{Proposal propagation}
After positional encoding by YOLOX proposals $P_0$ in the first frame, proposal queries are further refined by self-attention and interact with image features through deformable attention to produce track queries $q_{0}^{tr}$ and relative offsets $(\Delta x, \Delta y, \Delta w, \Delta h)$. The final prediction $Y_0$ is obtained by combining the proposals $P_0$ with the predicted offsets.

In subsequent frames, the object queries $\mathbf{q}_{t-1}^{tr}$ and predictions from the previous frame are propagated and concatenated with the proposals from the new frame $P_t$.

\subsection{Spatial-aware Sparse Memory}
\label{sec:method_SASM}
In this study, our objective is to optimize memory utilization in video processing by minimizing redundancy in object information storage. We propose a sparse storage strategy tailored to capture significant object deformations between frames, thereby avoiding unnecessary data retention during periods of minimal change. Acknowledging the diverse deformation characteristics across objects, a uniform storage approach proves inefficient. Consequently, we hypothesize a direct relationship between an object's movement dynamics and its deformation profile in video sequences, motivating an adaptive memory management scheme.

Our approach leverages the displacement of each object's centroid as a decisive criterion for selective memory storage. By dynamically adjusting storage intervals based on real-time deformation analysis, our method aims to enhance storage efficiency while minimizing redundancy information. Figure \ref{fig:SASMmot} delineates the architecture of our proposed memory management system, encompassing the following methodological steps:

\begin{figure}[!t]
    \centering
    \includegraphics[width=\linewidth]{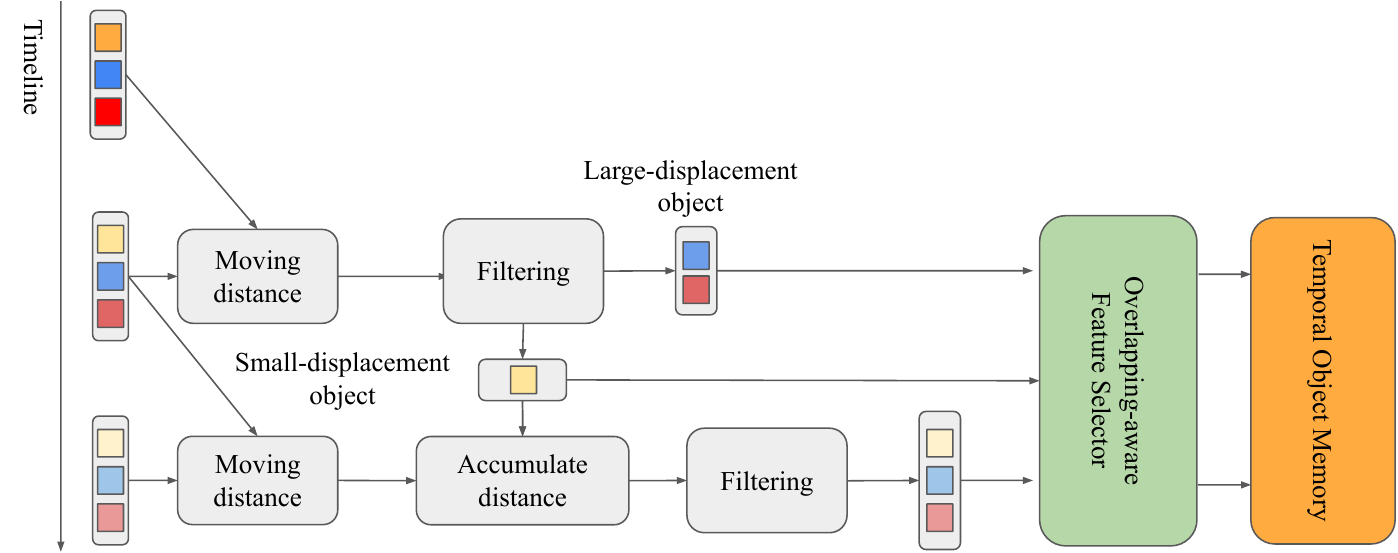}
    \caption{The Spatial-aware Sparse Memory (SASM) takes the output of MOTRv2 as input, which includes both object queries and coordinates. From the object coordinates, the module filters out small-displacement objects to accumulate distance to the next frame, and stores only objects with large displacement, as described in Section \ref{sec:method_SASM}. The Overlapping-aware Feature Selector (OFS) plays a crucial role in selecting the best features of each instance since the last memory update, as described in Section \ref{sec:method_ofs}.}    \label{fig:SASMmot}
\end{figure}

\begin{enumerate}
    \item \textbf{Calculating Moving Distance}: In frame \( t \), the moving distance \( d_{i, t} \) of object \( i \) is determined by the Euclidean distance between its centers in frames \( t \) and \( t-1 \). In particular, this distance is computed as the \( ||\cdot||_2 \) norm of the vector difference between the centers \( \mathbf{c}_{i, t} \) and \( \mathbf{c}_{i, t-1} \), respectively, where \( \mathbf{c}_{i, t} \) denotes the center coordinates of the bounding box for object \( i \) in frame \( t \), extracted from \( Y_t \):
    \begin{equation} \label{eq:cal_d} 
        d_{i, t} = ||\mathbf{c}_{i, t} - \mathbf{c}_{i, t-1}||_2
    \end{equation}
      
    \item \textbf{Accumulating Moving Distance}: If the moving distance \( d_{i, t} \) does not surpass $\epsilon$, it is accumulated with the moving distance of the object in the next frame \( t+1 \). This accumulation accounts for gradual movement across consecutive frames.
    \[
    d_{i, t+1} = d_{i, t+1} + d_{i, t}
    \]
    Otherwise, if the computed moving distance \( d_{i, t} \) exceeds a predefined threshold \(\epsilon\), indicating significant motion, the object's features are stored in memory. This step ensures that only objects undergoing substantial movement are considered for memory storage. Additionally, with a higher \(\epsilon\), longer temporal information of the object can be stored, but it may also lead to the omission of some important features.

    \item \textbf{Updating Object Query Feature}: The new object query feature \( \mathbf{q}_{i, t+1} \) for frame \( t+1 \) is computed as a weighted average of the previous query feature \( \mathbf{q}_{i, t} \) and the average query features \( \mathbf{q}_{i, j} \) of the \(i\)-th object stored in memory. This update mechanism incorporates both the current state of the object and historical information stored in memory, ensuring adaptability to changes in object appearance and position over time.
    \begin{equation}\label{eq:update_next_q}
    \mathbf{q}_{i, t+1} = \alpha \cdot \mathbf{q}_{i, t}^{tr} + \frac{(1 - \alpha)}{m} \sum_{j} \mathbf{q}_{i, j}^{tr}
    \end{equation}
    where \( j \) indexes frames in the memory, \( m \) is the number of frames in the memory, and \( \alpha \) is a hyperparameter determining the bias towards the current feature or the feature in memory.
\end{enumerate}

These steps collectively form a coherent framework for selective features stored in memory, leveraging object movement as a key criterion for memory allocation and ensuring the diversity of object state in object tracking across frames. As a result, this sparse memory strategy not only addresses the issue of bias towards the latest appearances of objects in feature representation but also facilitates the capture of longer temporal information by reducing redundancy when compared with dense memory. 

\subsection{Overlapping-aware Feature Selector}
\label{sec:method_ofs}
Empirical observations reveal that object features are often incorrectly stored during occlusion or overlap, introducing noise. To address this, we propose selectively storing object features in memory.

A simplistic approach of waiting for unobstructed views leads to significant temporal gaps. Instead, we select the feature from the frame with the lowest Intersection over Union (IoU) with other objects since the last update. This minimizes erroneous features and ensures smoother continuity.

We define the stored feature \( \mathbf{\hat{q}}_{i, t} \) at frame \( t \) as:

\begin{equation}
\label{eq:update_q}
\mathbf{\hat{q}}_{i, t}^{tr} = \begin{cases} 
\mathbf{q}_{i, t}^{tr} & \text{if } \text{IoU}_{i, t} < \text{IoU}_{i, k} \\
\mathbf{q}_{i, k}^{tr} & \text{otherwise} 
\end{cases}
\end{equation}

where \( \mathbf{q}_{i, t}^{tr} \) is the feature of object \( i \) at frame \( t \), and \( \mathbf{q}_{i, k}^{tr} \) is the feature at frame \( k \) with the highest IoU. This selection process retains more reliable features, reducing noise and e



As a result, this selective feature storage strategy contributes to our tracking system's overall efficiency and effectiveness, enabling more accurate and reliable object tracking over extended durations. 


\section{Experimental Results}
\subsection{Dataset and Metric}
In this paper, we evaluate our methods on the DanceTrack \cite{dancetrack} dataset. This dataset requires tracking multiple objects with uniform appearance and diverse motion.
To ensure fair comparisons, we assess our methods using well-known metrics as employed in previous works \cite{motr, motrv2}: HOTA \cite{hota} to evaluate our method and analyze the contribution decomposed into detection accuracy (DetA) and association accuracy (AssA); MOTA \cite{clearmot}; IDF1 \cite{id-metrics} metrics.


\subsection{Implementation Details}
This study leverages the power of MOTRv2 \cite{motrv2}, a model jointly trained on the CrowdHuman \cite{crowdhuman} and DanceTrack \cite{dancetrack} datasets, for object tracking. We utilize pre-trained weights from MOTRv2 and employ YOLOX \cite{yolox2021} for object proposal generation, using the publicly available pre-trained model from the DanceTrack GitHub repository\footnote{https://github.com/DanceTrack/DanceTrack}. Other MOTRv2 settings remain unchanged.

For the Spatial-aware Sparse Memory (SASM), we utilize two hyperparameters: the maximum number of stored objects (capped at 10) and the moving distance threshold (set to 0.1 of the image size), which were determined through experiments detailed in Section \ref{sec:exp_hyper}.


\subsection{Comparing with SOTA}
To demonstrate the effectiveness of our proposed method, we compared our performance with existing state-of-the-art (SOTA) methods on the DanceTrack dataset represented in Table \ref{tab:methods_comparison}. Our method's ability to leverage sparse memory for enhanced object re-identification and tracking consistency during object deformation drives this achievement. As shown in Table \ref{tab:methods_comparison}, our approach achieves significant improvements over MOTRv2, with gains of \textit{2.0 and 2.1 points, respectively, in AssA and IDF1 metrics}. These improvements translate to a HOTA score of 71.2, which convincingly underscores the efficacy of our memory-based strategy. Furthermore, despite employing the same object proposals, our method exhibits a marginal but notable DetA improvement of 0.3 compared to MOTRv2. This enhancement can be attributed to our method's superior object-tracking consistency.

\begin{table}[h]
\centering
\caption{Performance comparison of SOTA methods.}
\label{tab:methods_comparison}
\begin{tabular}{l|ccccc}
\toprule
\toprule
Methods & HOTA & DetA & AssA & MOTA & IDF1 \\
\midrule
CenterTrack \cite{centertrack} & 41.8 & 78.1 & 22.6 & 86.8 & 35.7 \\
TransTrack \cite{transtrack} & 45.5 & 75.9 & 27.5 & 88.4 & 45.2 \\
ByteTrack \cite{bytetrack} & 47.7 & 71.0 & 32.1 & 89.6 & 53.9 \\
QDTrack \cite{qdtrack} & 54.2 & 80.1 & 36.8 & 87.7 & 50.4 \\
MOTR \cite{motr} & 54.2 & 73.5 & 40.2 & 79.7 & 51.5 \\
OC-SORT \cite{ocsort} & 55.1 & 80.3 & 38.3 & 92.0 & 54.6 \\
MeMOTR \cite{memotr} & 68.5 & 80.5 & 58.4 & 89.9 & 71.2 \\
MOTRv2 \cite{motrv2} & 69.9 & 83.0 & 59.0 & 91.9 & 71.7 \\
MOTRv3 \cite{motrv3} & 70.4 & \textbf{83.8} & 59.3 & \textbf{92.9} & 72.3 \\
\midrule
\textbf{Ours} & \textbf{71.2} & 83.3 & \textbf{61.0} & 92.0 & \textbf{73.8} \\
\bottomrule
\bottomrule
\end{tabular}
\end{table}

\subsection{Ablation study}
To comprehensively evaluate the efficacy of our proposed method, we conduct experiments across three critical dimensions: assessing the effectiveness of each module, exploring alternative design choices for these modules, and rigorously selecting hyperparameters. To ensure fairness in comparison, we perform an ablation study on the validation set to identify the optimal design choices and hyperparameters.

\subsubsection{Effectiveness of each module} 

To initiate our experimentation, we systematically incorporated each module into the baseline model, MOTRv2 \cite{motrv2}, in an iterative manner, leading to the results depicted in Table \ref{tab:abl_1}. Our analysis indicates that integrating the Spatial-aware Sparse Memory (SASM), as elaborated in Section \ref{sec:method_SASM}, improves the IDF1 score by 1.0 and the AssA score by 1.0. Nevertheless, we observed performance degradation in specific videos due to noise stemming from object feature overlap within the stored memory. To tackle this issue, we introduce the Overlapping-aware Feature Selector (OFS) detailed in Section \ref{sec:method_ofs}. This addition substantially improves tracking consistency by enhancing both detection accuracy (DetA by 0.3) and tracking accuracy (AssA by 0.4).

\begin{table}[h]
\centering
\caption{The effectiveness of each proposed module}
\label{tab:abl_1}
\begin{tabular}{l|ccccc}
\toprule
\toprule
Methods & HOTA ($\uparrow$) & DetA ($\uparrow$) & AssA ($\uparrow$) & IDF1 ($\uparrow$) \\
\midrule
Baseline & 65.4 & 79.0 & 54.4 & 67.1 \\
+ SASM & 66.1 (\textcolor{green}{+0.6}) & 79.1 (\textcolor{green}{+0.1}) & 55.4 (\textcolor{green}{+1.0}) & 69.4 (\textcolor{green}{+1.0}) \\
+ OFS & 66.5 (\textcolor{green}{+0.4}) & 79.4 (\textcolor{green}{+0.3}) & 55.8 (\textcolor{green}{+0.4}) & 69.5 (\textcolor{green}{+0.1}) \\
\bottomrule
\bottomrule
\end{tabular}
\end{table}

\subsubsection{Alternative design comparison}
Our work introduces two memory modules: the Spatial-aware Sparse Memory (SASM) and the Overlapping-aware Feature Selector (OFS). We conducted a series of empirical studies to optimize each module's design. The first investigation focused on memory density, comparing feature storage at every frame (dense memory) with a sparse approach (Table \ref{tab:alb_design_comparision}). In both experiments, the maximum stored features were capped at 10. In particular, dense memory exhibited significantly lower performance across all metrics in the DanceTrack validation set. We attribute this disparity to rapid object deformation. For stationary objects, dense memory features tend to bias towards the current state, often missing sudden deformations common in the DanceTrack dataset. In contrast, our method stores fewer features during minor object deformations while preserving longer temporal information, aiding in more accurate object state associations.


Our next investigation focused on addressing the challenge of handling overlapping object features during storage. As described in Section \ref{sec:method_ofs}, we explored two approaches: delaying storage until encountering a frame with optimal object positioning or storing the feature at the best location since the last storage. Our findings uncovered a significant drawback with the delay approach: large gaps between feature storage events due to frequent object overlaps (Table \ref{tab:alb_design_comparision}). Both experiments used the SASM module to ensure a fair comparison.

\begin{table}[t!]
\centering
\caption{Comparison between Memory Types and Different Strategies}
\label{tab:alb_design_comparision}
\begin{tabular}{l|cccc}
\toprule
\toprule
\multirow{2}{*}{Design} & \multicolumn{4}{c}{Performance Metrics ($\uparrow$)} \\
\cline{2-5}
& HOTA & DetA & AssA & IDF1 \\
\midrule
Dense & 64.5 & 78.3 & 53.3 & 67.2 \\
Sparse (Ours) & 66.1 (\textcolor{green}{+1.6}) & 79.1 (\textcolor{green}{+0.8}) & 55.4 (\textcolor{green}{+2.1}) & 69.4 (\textcolor{green}{+2.2})  \\
\midrule
\midrule
Delaying & 66.0 & 79.0 & 55.3  & 68.8 \\
OFS (Ours) & 66.5 (\textcolor{green}{+0.5}) & 79.4 (\textcolor{green}{+0.4}) & 55.8 (\textcolor{green}{+0.5}) & 69.5 (\textcolor{green}{+0.7}) \\
\bottomrule
\bottomrule
\end{tabular}
\end{table}



\subsubsection{Hyperparameters}
\label{sec:exp_hyper}
We conducted thorough experiments to explore our algorithm's sensitivity to hyperparameters. Initially, we focused on determining the optimal number of frames to store in memory for each object. Our investigation revealed that utilizing 10 frames yielded the most favorable outcomes. Conversely, as detailed in Table \ref{tab:merged_comparison_vertical}, alternative configurations resulted in notably inferior performance. This discrepancy arises from the inherent challenge of aligning memory features with object features, a task complicated by the absence of explicit training in our algorithm. The more extended number of frames ($\geq 15$ frames) of memory exacerbates this issue by introducing sparse data and noise from tracking failures, leading to a substantial mismatch between memory and object features. Conversely, a paucity of frames (5 frames) constrains the algorithm's capacity to capture sufficient temporal information.

We also evaluated the impact of adjusting the threshold on our algorithm's performance (Table \ref{tab:merged_comparison_vertical}). Setting the threshold ($\epsilon$) to 0.05 aligns closely aligns the algorithm's performance with the baseline. However, thresholds exceeding 0.1 relative to image size result in prolonged movement distances, delaying feature storage in memory. This exacerbates disparities between stored and current object features, leading to a performance decline to around 65.8\%. Nonetheless, even with highly sparse memory, our algorithm shows performance improvements compared to the baseline, achieving 65.4\%.


\begin{table}[!t]
\centering
\caption{Comparison between different parameters: moving threshold vs memory length}
\label{tab:merged_comparison_vertical}
\begin{tabular}{c|cccc}
\toprule
\toprule
\multirow{2}{*}{Parameter} & \multicolumn{4}{c}{Performance Metrics ($\uparrow$)} \\
\cline{2-5}
& HOTA & DetA & AssA & IDF1 \\
\midrule
Moving Threshold ($\epsilon$) \\
\midrule
0.05 & 65.5 & 79.1 & 54.5 & 68.7 \\ 
0.1 & \textbf{66.5} & \textbf{79.4} & \textbf{55.8} & \textbf{69.5} \\ 
0.2 & 65.7 & 79.1 & 54.8 & 68.5 \\ 
0.3 & 65.9 & 79.2 & 55.1 & 68.7 \\ 
0.4 & 65.8 & 79.0 & 55.0 & 68.5 \\ 
\midrule
Memory Length (\# features) \\
\midrule
5  & 66.0 & 79.1 & 55.2 & 68.9 \\
10 & \textbf{66.5} & \textbf{79.4} & \textbf{55.8} & \textbf{69.5} \\
15 & 65.8 & 79.1 & 54.9 & 68.5 \\
20 & 65.6 & 78.6 & 54.9 & 68.6 \\
\bottomrule
\bottomrule
\end{tabular}
\end{table}

The ablation study on hyperparameters underscores a critical issue inherent in the method: its sensitivity to hyperparameter settings. This sensitivity arises from the challenge of effectively managing the gap between memory features and current features.

\subsection{Qualitative result}
\begin{figure}[!t]
    \includegraphics[width=\linewidth,center]{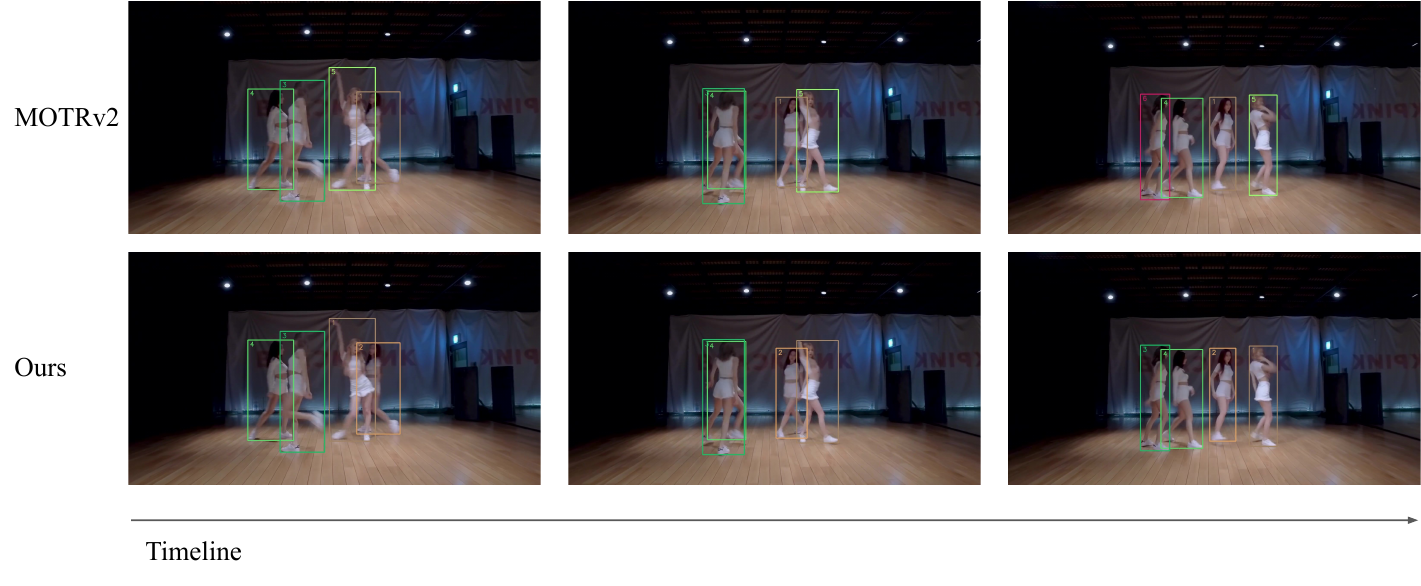}
    \caption{Qualitative results for tracking consistency}    
    \label{fig:qualitative-track-consistent}
\end{figure}

Our approach stands out for its proficiency in tracking objects across two crucial qualitative scenarios: maintaining consistency during changes in viewpoint and re-identifying them after missed detections. As depicted in Figure \ref{fig:qualitative-track-consistent}, consider a person (ID 3) whose appearance alters due to a body rotation. In such instances, MOTRv2 faces challenges, often assigning a new ID when the person turns around. Conversely, our method capitalizes on its extensive temporal memory, ensuring the retention of the object's original ID.

\section{Conclusion}
In conclusion, by introducing a novel memory mechanism, our study builds upon the previous MOTRv2 framework, a hybrid architecture for Multiple Object Tracking (MOT) problems. Specifically, we focus on optimizing feature selection for object memory within the MOT context, which often involves numerous objects and quickly escalates in complexity with dense memory. Unlike prior approaches, we propose a sparse memory scheme based on object movement.

This study presents an effective method for sparsely storing object features based on their movement and overlapping patterns. As demonstrated through experiments, this results in significant performance enhancements over baseline methods. Additionally, we thoroughly evaluate each model design to glean deeper insights into this approach.

However, despite the benefits of our training-free approach, we acknowledge that the sensitivity of memory object features to hyperparameters can pose challenges in aligning them with the MOTRv2 pipeline. This issue warrants further investigation to design adaptors that can seamlessly integrate the features of each module into an end-to-end model architecture.
\section*{Acknowledgements}
\label{sec:ackn}
This research is funded by University of Science, VNU-HCM under grant number T2023-93.
Furthermore, Nguyen Quang Thuc was also funded by the Master, PhD Scholarship Programme of Vingroup Innovation Foundation (VINIF), code VINIF.2022.ThS.JVN.10.

\printbibliography
\end{document}